\documentclass[10pt,twocolumn,letterpaper]{article}

\usepackage{cvpr}              
\usepackage[pagebackref,breaklinks,colorlinks]{hyperref}
\usepackage{amsthm}
\usepackage{amsmath}
\usepackage{multirow}
\usepackage{booktabs} 
\usepackage{graphicx}
\usepackage{lipsum}
\usepackage[utf8]{inputenc}

\usepackage[pagebackref,breaklinks,colorlinks]{hyperref}

\def\paperID{298} 
\def\confName{3DV\xspace}
\def\confYear{2024\xspace}

\title{UAVD4L: A Large-Scale Dataset for UAV 6-DoF Localization}

\author{Rouwan Wu$^{1*}$  
\quad Xiaoya Cheng$^{1*}$  
\quad Juelin Zhu$^{1}$ 
\quad Xuxiang Liu$^{1}$ 
\quad Maojun Zhang$^{1\dagger}$\\[1.5mm] 
\quad Shen Yan$^{1}$  \\
 $^1$National University of Defense Technology\quad \\
ChangSha, China\\
{\tt\small \{wurouwan97, chengxy, zhujuelin, liuyuxiang17, mjzhang, yanshen12\}@nudt.edu}
}

\begin{document}
\twocolumn[{%
\renewcommand\twocolumn[1][]{#1}%
\maketitle
\begin{center}
    \centering
    \captionsetup{type=figure}
\includegraphics[width=0.9\textwidth,height=9cm]{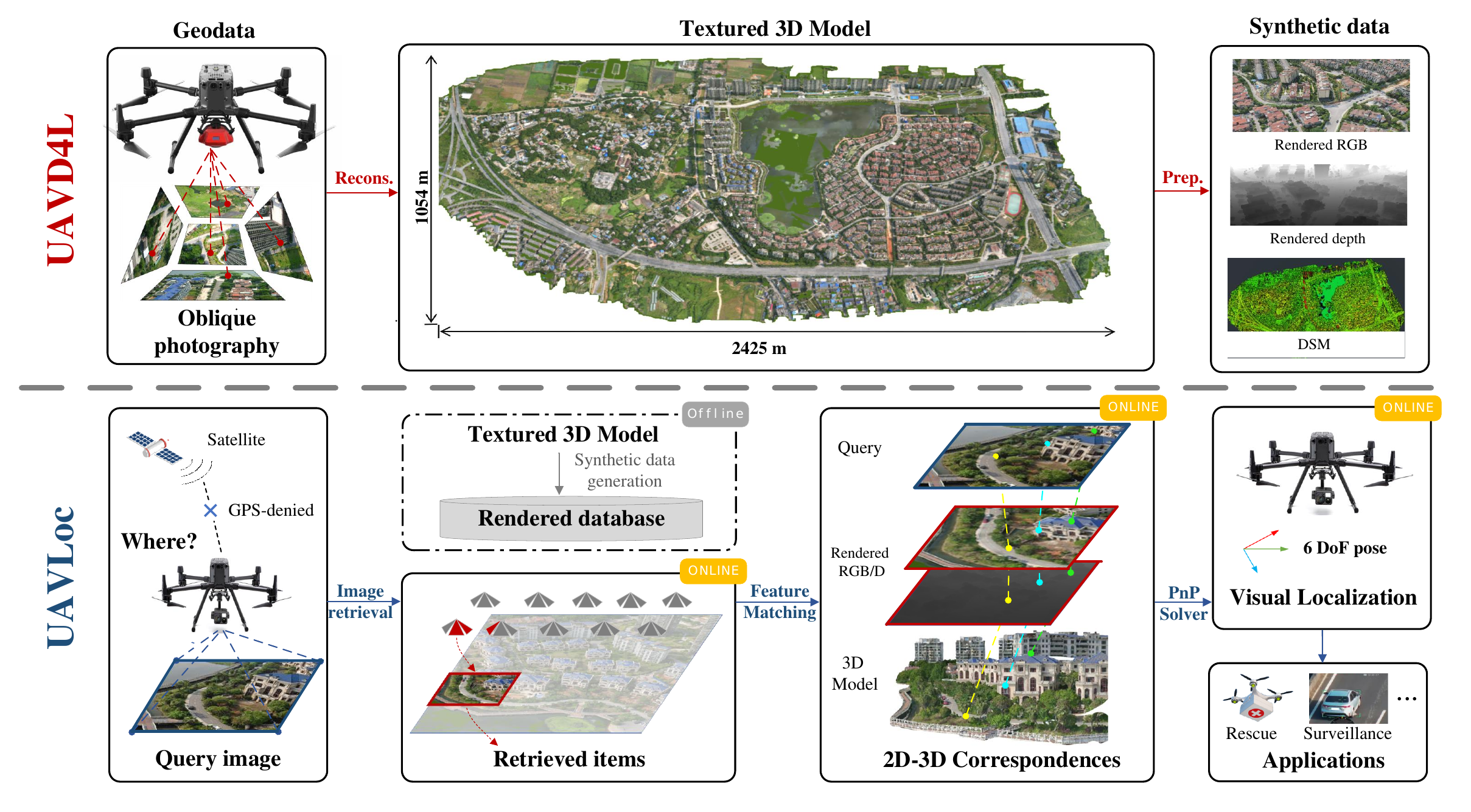}
    \captionof{figure}{\textbf{Top.} We introduce a large-scale dataset for the 6-DoF localization of UAVs. The dataset includes a 3D reference textured model, which enables the generation of synthetic data such as rendered RGB and depth images, as well as a Digital Surface Model (DSM). \textbf{Bottom.} We also develop an offline-and-online pipeline for performing 6-DoF localization of UAVs in GPS-denied environments.}
    \label{main_figure}
\end{center}%
}]
 \let\thefootnote\relax\footnotetext{$^\dagger$Corresponding author: Maojun Zhang.}
 \let\thefootnote\relax\footnotetext{$^*$Equal contribution.}
 
\begin{abstract}
Despite significant progress in global localization of Unmanned Aerial Vehicles (UAVs) in GPS-denied environments, existing methods remain constrained by the availability of datasets. Current datasets often focus on small-scale scenes and lack viewpoint variability, accurate ground truth (GT) pose, and UAV build-in sensor data. To address these limitations, we introduce a large-scale 6-DoF UAV dataset for localization (UAVD4L) and develop a two-stage 6-DoF localization pipeline (UAVLoc), which consists of offline synthetic data generation and online visual localization. Additionally, based on the 6-DoF estimator, we design a hierarchical system for tracking ground target in 3D space. Experimental results on the new dataset demonstrate the effectiveness of the proposed approach. Code and dataset are available at \href{https://github.com/RingoWRW/UAVD4L}{https://github.com/RingoWRW/UAVD4L}.
\end{abstract}
\section{Introduction}
\label{sec:intro}
Global localization of Unmanned Aerial Vehicles (UAVs) plays an important role in various applications such as cargo transport~\cite{larson2021autonomous}, surveillance~\cite{guisado20193d}, and search-rescue tasks~\cite{silvagni2017multipurpose,bejiga2017convolutional}. 
While most established works rely on global satellite navigation systems (GNSS)~\cite{couturier2021review,lu2018survey,balamurugan2016survey} for global position, vulnerabilities in GNSS signal reception~\cite{kanellakis2017survey,al2018survey,xu2018vision} have led to the development of visual solutions~\cite{nassar2020aerial} for GPS-denied environments.

However, compared with ground-level localization of cellphones or VR devices~\cite{sarlin2019coarse,panek2022meshloc,sarlin2021back}, the development of visual localization of UAVs lags behind.
We believe this is partly due to the scarcity of open-source datasets for localizing airborne platforms in the academic community.
Among the most evaluated open-source datasets, limitations include restricted to small scenes~\cite{cisneros2022alto}, evaluation of only 3-DoF positions~\cite{zheng2020university}, limited viewpoint variability~\cite{goforth2019gps}, lack of accurate ground truth (GT)~\cite{gurgu2022vision}, concentration on learning-based regression methods~\cite{yan2022crossloc}, and disregard for additional inertial sensors~\cite{marcu2018multi}.

To facilitate research in this area, our first contribution is to introduce a \textbf{novel large-scale dataset for 6-DoF UAV localization}, as shown in the top part of Figure~\ref{main_figure}. The dataset comprises a textured 3D reference model reconstructed from aerial oblique photography, covering approximately 2.5 million square meters. It enables the generation of various synthetic data, including rendered RGB and depth images, and a Digital Surface Map (DSM).
Query images are captured and sampled from five distinct flight trajectories, with varying heights (50-200 meters), viewpoints (pitch angles of 15-70 degrees), and acquisition positions. Rather than relying on built-in sensor information from the UAV, such as Global Positioning System(GPS) or Real-time kinematic(RTK), to provide 3-DoF ground truth positions, we use manual tie points to register query sequences to the reference map for more accurate 6-DoF ground truth poses.
In addition, sensor data (i.e., the rotation priors from the Inertial Measurement Unit (IMU)) is recorded to assist further localization algorithms.

Second, we develop a new \textbf{two-stages 6-DoF UAV localization pipeline for GPS-denied environments}, which consists of offline synthetic data generation and online visual localization, as illustrated in the bottom portion of Figure~\ref{main_figure}. 
In the offline stage, our approach employs rendering techniques to obtain synthetic views and depth maps at different virtual viewpoints (with varying heights and directions) to fully represent the dense 3D model of the scene.
In the online stage, inspired by the recent ground-level visual localization method, SensLoc~\cite{yan2023long}, we combine camera information with prior data from UAV equipment. 
Specifically, we first use rotation information from the build-in IMU to constrain search space for identifying relevant reference views of the query image. Then, feature matches between the query image and the top-$k$ retrieved database images are established and lifted to 2D-3D correspondences  with the help of depthmaps. Finally, a gravity-guided PnP RANSAC~~\cite{cai2022review,huang2020high} is employed to estimate the camera pose.

Finally, based on the 6-DoF localization results of UAVs, we design a \textbf{hierarchical system to track designated objects located on the ground}, such as pedestrians or vehicles.
This system employs a wide-angle lens (offering a larger field of view) to determine the 6-DoF pose of the UAV, and a zoom lens to identify targets, facilitating more accurate target extraction on the 2D image. 
A ray-tracing technique~\cite{huang2020high} is utilized to project 2D targets on image plane onto their absolute positions in the 3D map.

\begin{table*}[]
\resizebox{\textwidth}{!}
{
\begin{tabular}{ccccccc}
\toprule
Name & Reference source & UAV viewpoint & Map size &  Estimated pose DoF & Ground truth & Open source \\
\midrule
Chen et al.~\cite{chen2021real} & Google Earth & arbitrary & $400 \times 400 m^2$ &  6-DoF & GPS and IMU & no \\
Goforth et al.~\cite{goforth2019gps} & United States Earth & top view & $0.85km$ &  6-DoF & RTK-GPS and IMU & yes \\
Kinnari et al.~\cite{kinnari2021gnss} & georeferenced orthophotos & arbitrary & $200 \times 200 m^2$ &  6-DoF & RTK-GPS and IMU & no \\
Kinnari et al.~\cite{kinnari2022lsvl} & Google Earth & arbitrary & - &  3-DoF (x,y,heading) & RTK-GPS and IMU & no \\
Patel et al.~\cite{patel2020visual} & Google Earth & top view & $300 \times 350 m^2$ & 4-DoF (x,y,z,heading) & RTK-GPS and IMU & no \\
Gurgu et al.~\cite{gurgu2022vision} & Google Earth & top view & - & 2-DoF (x,y) & RTK-GPS & yes \\
Zheng et al.~\cite{zheng2020university} & Google Earth and Google Map& arbitrary & - &  2-DoF (x,y) & sim from Google Earth & yes \\
Yan et al.~\cite{yan2022crossloc} & render and real geographic image & arbitrary & 2.7 million $m^2$ &  6-DoF & RTK-GPS and ground control point & yes \\
Cisneros et al.~\cite{cisneros2022alto} & LiDAR, real image & arbitrary & $50 \times 50 m^2$ & 6-DoF & RTK-GPS and ground control point & yes \\
\cline{1-7}
Ours & 3D model, geographic image, DSM & arbitrary & 2.5 million $m^2$ &  6-DoF & RTK-GPS and ground control point & yes \\
 \bottomrule
\end{tabular}
}
\caption{\textbf{Overview of the existing UAV localization datasets.} Several key attributes are taken into account: the source of reference data, the viewpoint, degrees of freedom used to measure the estimated pose, how to acquire ground truth data, and whether or not the dataset is open source.}
\label{tab2.1}
\end{table*}

\section{Related Work}
\label{sec:Related work}

\subsection{Ground-level Visual Localization} 
Conventionally, most established methods~\cite{panek2022meshloc,sarlin2019coarse,sarlin2020superglue, yan2023long} perform ground-level visual localization by establishing 2D-3D correspondences between the query image and the reconstructed sparse point clouds. The camera pose is then recovered using a PnP solver~\cite{haralick1994review, lepetit2009ep} with RANSAC~\cite{fischler1981random}. To scale to large scenes, an image retrieval phase~\cite{arandjelovic2016netvlad, gordo2017end, ge2020self} is often used to identify a small set of potentially relevant database images.
Recently, HLoc~\cite{sarlin2019coarse,sarlin2020superglue} has provided a comprehensive toolbox that integrates existing methods and achieves a promising result.
To address the robustness, efficiency, and flexibility issues of HLoc, subsequent research includes replacing the sparse point clouds with dense model~\cite{panek2022meshloc}, and introducing sensor prior information~\cite{sarlin2022lamar, yan2023long}.

Inspired by these works, we propose a two-stage pipeline that effectively explores synthetic view representation of the dense model and successfully combines sensor prior with visual signal for localizing the 6-DoF pose of UAVs.

\subsection{GPS-denied UAV Localization}
Compared with ground-level visual localization, pose estimation of UAVs in GPS-denied environments has received less attention. 
Some previous works~\cite{xu2022uav,wu2021coarse,yin2023isimloc,yin2021i3dloc} treat UAV localization as a scene recognition task, where the goal is to identify the most similar images to a query image within a geo-tagged database.
The performance of this approach is influenced by the density and distribution of images within the database and can result in significant errors.
Other works~\cite{nath2022drone,yan2022crossloc,bianchi2021uav,choi2020brm,liu2023uav} aim to calculate the 6-DoF camera pose and can be broadly classified into two categories.
The first category~\cite{yan2022crossloc,shetty2019uav} involves using neural networks to implicitly encode the scene and regress an absolute pose for an input image. These methods have the advantage of occupying small memory and having fast inference time, but their localization accuracy is limited.
The second category~\cite{chen2021real} follows HLoc~\cite{sarlin2019coarse, sarlin2020superglue} to explicitly use the map to render synthentic views upon Google Earth and recovers 6-DoF pose via image retrieval, feature matching and PnP RANSAC.
Our work is similar to the second one, but we explore richer rendering viewpoints to represent the scene, and make full use of sensor priors to achieve more accurate 6-DoF localization of UAVs.

\subsection{UAV Localization Datasets} 
The development of UAV localization research is largely driven by the availability of datasets, as summarized in Table~\ref{tab2.1}. However, existing UAV datasets face several challenges that limit their effectiveness in advancing the field. These challenges include: 
1) \textbf{Limited Degrees of Freedom.} Many datasets, such as University-1652~\cite{zheng2020university} and ALTO~\cite{cisneros2022alto}, provide only 3 degrees of freedom (DoF) for localization, which may not be sufficient for certain applications.
2) \textbf{Top-View Bias.} Some datasets, such as those presented in~\cite{goforth2019gps} and~\cite{gurgu2022vision}, focus exclusively on top-view images generated from Google Earth or real drones. However, these datasets do not represent the diverse scenarios of real-world UAV usage. 
3) \textbf{Regression-Based Approaches.} Datasets such as CrossLoc~\cite{yan2022crossloc} provide rich data sources including real images, synthetic images, and semantic information for regression-based localization approaches. However, these approaches may not be suitable for structure-based localization algorithms.

These challenges highlight the need for more comprehensive and diverse datasets to advance the field of UAV localization.

\section{Dataset}
\label{sec:dataset}
\begin{figure}[htbp]
    \centering
  \includegraphics[width=0.45\textwidth]{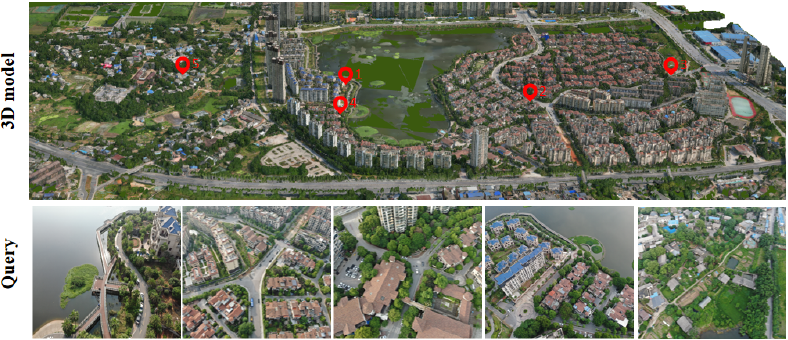}
    \caption{\textbf{Distribution of query images.} The query sequence consists of five trajectories, with the first four covering urban areas characterized by a high density of buildings, while the fifth trajectory covers a rural area with predominantly vegetation. The red locators, numbered from 1 to 5, represent the shooting positions.}
    \label{dataset_vis}
\end{figure}

We present a large-scale dataset, UAVD4L, that covers an area of approximately 2.5 million square meters and includes a diverse range of urban and rural scenes, including buildings, streets, vegetation, and a lake. A visualization of the dataset is provided in Figure~\ref{dataset_vis}.

\subsection{Reference Map Collection}
To create the reference map, we captured high-resolution aerial images $\{I_o\}$ using a DJI M300 RTK drone\footnote{https://enterprise.dji.com/cn/matrice-300} equipped with a five-eye oblique SHARE PSDK 102s camera\footnote{https://shareuav.cn/V3S}. 
To ensure sufficient and uniform coverage of the area, we pre-planned a grid fashion flight path, which was automatically executed by the drone’s flight control system. We then applied modern 3D reconstruction techniques to produce textured mesh models and align them with the real geographic coordinate system using built-in RTK measurements and ground control points.
More details about reference map collection is provided in the supplementary materials.

\subsection{Query Image Collection}
We collect all query images $\{I_q\}$ using a DJI M300 RTK mounted with a DJI H20T camera\footnote{https://enterprise.dji.com/cn/zenmuse-h20-series}.
To enhance the diversity of query images, we manually captured images cross different regions, flight altitudes and a diverse range of capturing angles. 
In addition, to simulate the different shooting habits of drone pilots, we captured query images using two modes: continuous shooting at 2-second intervals and random manual shooting. The supplementary material provides statistics on the number of query images for each trajectory.

\subsection{Query GT Generation}
In this work, we adopt a scalable semi-automatic annotation method to generate pseudo GT poses for query images. 
Our method is capable of producing pose annotations for hundreds of query images with minimal human intervention. 
The main steps of the method are as follows: First, we use the Structure-from-Motion (SfM) method to separately reconstruct sparse point clouds from the oblique photographic image $\{I_o\}$ and multiple query sequences $\{I_q\}$. Next, we manually select some tie-points between query sequence and the oblique photographic images and perform registration based on these tie-points. Finally, we refine SfM block using bundle adjustment to achieve a whole 3D model with oblique photographic and query images.

\begin{figure}[htbp]
    \centering
  \includegraphics[width=0.45\textwidth]{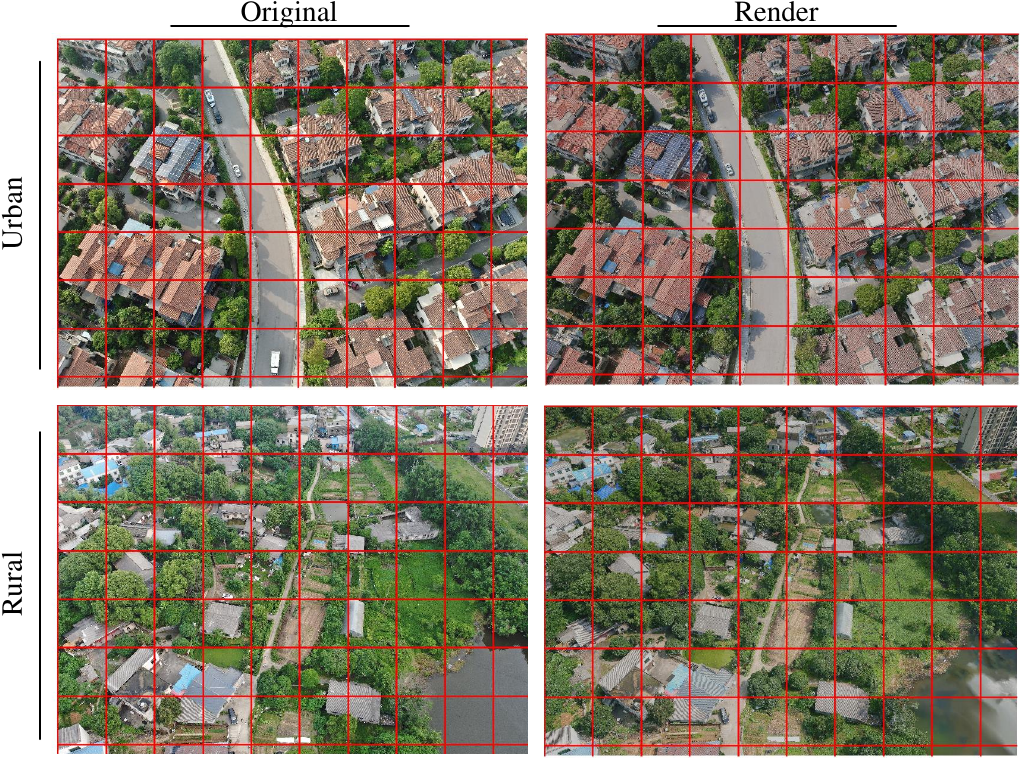}
    \caption{\textbf{GT poses quality on UAVD4L.} Pixel-aligned renderings of the estimated camera pose confirm that the poses are sufficiently accurate for evaluation.}
    \label{gt_render}
\end{figure} 

To evaluate the accuracy of the reference 3D model $\mathcal{M}$, we report the median reprojection error and the median reprojection error per tie-point, which are 0.82 pixels and 0.62 pixels, respectively. 
Besides, to ensure the accuracy of the pseudo GT pose, we conduct a visual inspection, and select the images whose pseudo GT pose rendered image is aligned with the original query image. Figure~\ref{gt_render} illuminates an example where the pixel-alignment between rendered and original query image indicates that the GT poses are accurate.
\section{Method}
\label{method}

\begin{figure*}[htbp]
    \centering
  \includegraphics[width=0.99\textwidth,height=8cm]{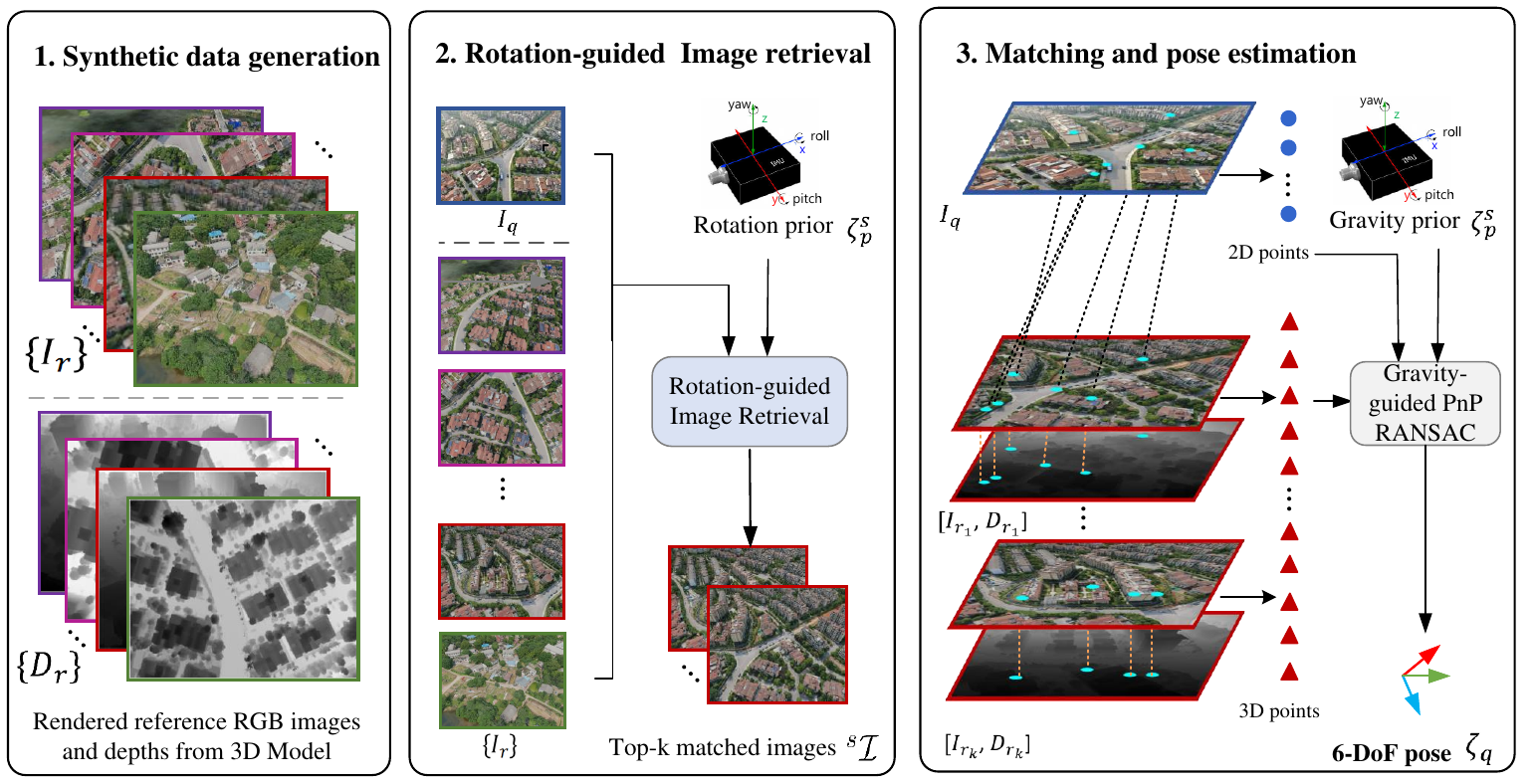}
    \caption{\textbf{Overview of the proposed method.} 1. We generate comprehensive synthentic data from textured 3D model, including RGB images ${I_r}$ and depthmaps ${D_r}$ (Section~\ref{database}). 2. For each query image $I_q$, we use an image retrieval algorithm combined with rotation sensor prior $S$ to find the top-$k$ relevent images. 3. Then, we apply feature detection and matching algorithm to establish the 2D-3D correspondence between the query image $I_q$ and the relevent images ${I_r}$. A gravity-guided PnP RANSAC is used to obtain the pose $\zeta_q$ of the UAV (Section~\ref{localization}). }
    \label{pipeline_fig}
\end{figure*}

Given a textured reference 3D model $\mathcal{M}$ with real-world geographical coordinates and the sensor information $S_{q}$ of query image $I_q$, our goal is to estimate the 6-Dof pose $\zeta _{q}$ of the query image $I_q$. 
To achieve this, we propose a two-stage localization pipeline.
In the first stage, the database images are rendered offline according to carefully selected virtual viewpoints (Section~\ref{database}). In the second stage, we conduct a 6-DoF localization pipeline online (Section~\ref{localization}). 
Figure~\ref{pipeline_fig} provides an overview of the proposed method. 
Furthermore, based on the results of UAV 6-DoF localization, we design a hierarchical system capable of tracking ground targets in 3D space (Section~\ref{target localization}).

\subsection{Synthetic Data Generation.}
\label{database}

Unlike ground-level localization situations, query image $I_q$ and oblique images $\{I_o \}$ are often captured at drastically varying positions and directions, posing a significant challenge for image retrieval~\cite{arandjelovic2016netvlad, gordo2017end, ge2020self} and feature matching algorithms~\cite{sun2021loftr,mughal2021assisting,barbarani2023local}.
To address this challenge, inspired by view synthesis work~\cite{moreau2022lens,zhang2021reference,zhang2022rendernet,taira2018inloc,yan2022crossloc}, we generate comprehensive synthetic reference data, including RGB images $\{I_r \}$ and depth maps $\{D_r \}$, to represent the dense map of the scene.

Specifically, we first leverage the geographical boundary of 3D map $\mathcal{M}$ to identify the area where view synthesis is required. Then the synthetic data generation can be divided into two parts: 1) For translation, we arrange the virtual viewpoints $\zeta$ above the 3D map and establish two hyperparameters for horizontal interval distance of $\alpha_t$ and vertical height of $H$, which are dynamically adjusted to ensure a uniform coverage of the entire geographic region. 2) For rotation, we set multiple $\theta_{pitch}$ and an appropriate interval $\alpha_{theta}$ of yaw to cover $360^\circ$ at the same position. Based multi view and multi height render method, it can enhance adaptation to the scale invariance and rotational invariance of query images.

\subsection{Visual Localization}
\label{localization}

\paragraph{Rotation-guided image retrieval.} 
Given a query image $I_q$, we aim to identify a set of commonly visible images $^{s}\mathcal{I} = \{I_{r1}, I_{r2},... I_{rk} \}$ within the reference images $\{I_r\}$, where $k$ represents the number of retrieved items. 
A general solution involves mapping the images $I_q, \{ I_r \}$ into a compact feature space using an embedding function $f(\cdot)$, followed by searching for the nearest neighbors of $I_q$ using the distance metric $d(f(I_q), f(I_r))$. 
However, simply using a global feature to find retrieved items may lead to some mistakes, as different UAV images share significant viewpoint and scale changes. 

To improve the retrieval accuracy, inspired by SensLoc~\cite{yan2023long}, we use the metadata from UAV sensors $S$, especially the rotation information, as a prior to pre-filter incorrect reference candidates.
Considering a richer variety of viewpoints in drones, we impose constraints on each degree of rotation angle, consisting of $\theta_{roll}$, $\theta_{yaw}$, $\theta_{pitch}$, as shown in Equation~\ref{equ3.1}.

\begin{equation}
\begin{split}
& \{ ||-\arcsin (R^{S}_{31}) - \theta^{r}_{roll}|| \le \gamma _o, \\
& ||\arctan (\frac{R^{S}_{21}}{R^{S}_{11}} ) - \theta^{r}_{yaw}|| \le \gamma _o, \\
& ||\arctan (\frac{R^{S}_{32}}{R^{S}_{33}}) -\theta^{r}_{pitch}||  \le \gamma _o, \\
& R^{S}_{31} \ne \pm 1 \},
\end{split}
\label{equ3.1}
\end{equation}

where $\gamma_o$ is a orientation threshold and $R^{S}$ represents the rotation matrix from the sensor prior $S$. 
$\theta^r$ represents the rotation angle of the reference image. 
After pre-filtering incorrect candidate data using rotation information as a prior, we are able to efficiently and accurately determine $k$ nearest neighbor $^{s}\mathcal{I}$ according to $d(f(I_q), f(^{p}\mathcal{I})).$

\paragraph{Matching and pose estimation.} We adopt learnable feature matching techniques to establish 2D-2D correspondences between the query image $I_q$ and the retrieved synthetic images $^s\mathcal{I}$.
Since depthmaps ${^s\mathcal{D}}$ are also rendered during the offline stage of synthetic data generation, these 2D-2D correspondences can be lifted to 2D-3D correspondences via back-projection.

For pose estimation, we follow SensLoc~\cite{yan2023long} and incorporate a gravity verification module into the PnP RANSAC process.
During each RANSAC iteration, we compute the deviation $d_{\epsilon}$ between the gravity direction of the sensor pose $\zeta ^{g}_s$ and the hypothetical pose $\zeta ^{g}_{hyp}$, as shown in Equation~\ref{equ3.2}.
If this deviation $d_{\epsilon}$ is less than a predefined stopping threshold $\gamma_{\epsilon}$, we halt the RANSAC iterations prematurely.

\begin{equation}
\label{equ3.2}
\begin{split}
    d_{\epsilon} = \arccos(\zeta ^{g}_s \cdot \zeta ^{g}_{hyp})
\end{split}
\end{equation}

\subsection{Hierarchical Target Tracking}
\label{target localization}

\begin{figure}[htbp]
    \centering
\includegraphics[width=0.49\textwidth]{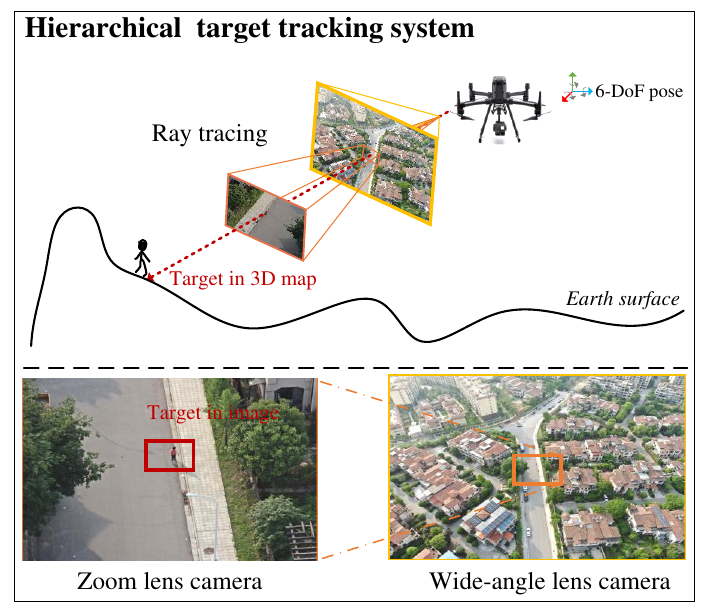}
    \caption{\textbf{The hierarchical target tracking system} consists of two lens: a wide-angle lens camera and a zoom lens camera. The wide-angle lens camera is used to recover the 6-DoF pose of the UAV, while the zoom lens camera is used to accurately detect the target. A ray-tracking technique is employed to track designated targets in 3D space.}
    \label{target}
\end{figure}

Based on the proposed 6-DoF UAV pose estimation approach mentioned above, we design a hierarchical system for target tracking on the ground, as illustrated in Figure~\ref{target}.

The system comprises two cameras: a fixed wide-angle lens camera and a zoom lens camera, which are carefully calibrated. 
The wide-angle lens, with a broader vision for global and local feature extraction, captures images for online 6-DoF localization. The zoom lens captures images for specific target detection, where the target occupies a sufficient area.
Afterwards, the 3D position of the target is estimated using ray-tracing~\cite{cai2022review,huang2020high} based on a digital elevation model (DEM), as detailed in the supplementary material. 

\subsection{Implementation Details}
In the offline synthetic reference data generation process, we generate virtual viewpoints with absolute altitude $H = 100m$ and $150m$, the horizontal interval $\alpha_{t}=50m$ and $\alpha_{t}=75m$, respectively, and generate 16 images with different angles ($\theta^{r}_{pitch}=0^\circ$ or $45^\circ$, $\alpha_{theta}=45^\circ$) for each same position. In the image retrieval step, the angle threshold $\gamma_o$ is set to $30^\circ$. In the gravity-guided PnP solver step, we set the gravity angle threshold $\gamma_{\epsilon}=2^\circ$. We perform all experiments on a PC equipped with Intel Core i9-11900K processor, GTX 3090 graphics card (24 GB RAM) and Ubuntu 18.04 operating system. We implement the proposed pipeline in Python alone with Pytorch.

\section{Experiment}
\label{sec:experiment}

\begin{figure*}[htbp]
    \centering
    \includegraphics[width=0.85\textwidth]{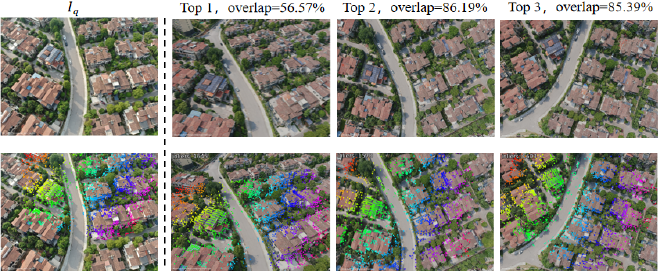}
    \caption{\textbf{Qualitative results.} The top part of the figure illustrates an example retrieval result using OpenIBL. The bottom part displays the local feature matching results by employing SuperPoint and SuperGlue. The matched correspondences are visualized using the same color scheme.}
    \label{retrieval_label}
\end{figure*} 

\subsection{Image Retrieval}
\label{5.1}

\paragraph{Baselines.} 
In this study, we evaluate the performance of several global descriptors at image retrieval stage, including NetVLAD~\cite{arandjelovic2016netvlad}, AP-GeM~\cite{gordo2017end}, and OpenIBL~\cite{ge2020self}. 
Additionally, we conduct controlled experiments to verify the effectiveness of using a rotation sensor prior.
\paragraph{Evaluation protocol.} 
Retrieval results are considered correct if they share a sufficient overlapping area with the query image, where the overlap percentage $P_{overlap}$ is greater than $50\%$.
The overlap area between two images is calculated by transforming pixels from one image to the other based on their depth maps. Please refer to the supplementary materials for more details.
\paragraph{Results.} The retrieval results are shown in Table \ref{tab4.1}. We maintain a rank list of size $k$ for each query and report recall and precision metrics. 
The experimental results show that all three representative retrieval methods are capable of finding correct retrieved items, even when the rotation sensor is not adapted. 
This can be attributed to the generation of comprehensive virtual views across the scene, which contributes to the availability of more visual similar images. 
Among all options, OpenIBL~\cite{ge2020self} with rank list $k=3$ and employing sensor priors leads the benchmark, and is chosen to find a visible reference set in the localization pipeline. 

\begin{table}[!ht]
\centering
\small
\begin{tabular}{lclll}
\toprule
Global feature    & Prior & R@1 & R@3 & P@3 \\ \hline
\multirow{2}{*}{NetVLAD~\cite{arandjelovic2016netvlad}} & - & 73.6 & 98.08 & 74.42 \\
                  & Rot & \textbf{79.9} & 98.08 & \textbf{77.8} \\
                  \hline
\multirow{2}{*}{AP-GeM~\cite{gordo2017end}} & - & 84.62 & 96.5 & 72.55 \\
                  & Rot & \textbf{85.31} & \textbf{97.55} & \textbf{73.14} \\
                  \hline
\multirow{2}{*}{OpenIBL~\cite{ge2020self}} & - & 83.74 & 98.43 & 76.69 \\
                  & Rot & \textbf{84.44} & \textbf{98.78} & \textbf{79.08} \\
\bottomrule
\end{tabular}
\caption{\textbf{Image retrieval results.} We report the top-$k$ recall and precision for $k=1,3$, using global image features and considering the use of rotation priors. }
\label{tab4.1}
\end{table}

\begin{figure}[htbp]
    \centering
  \includegraphics[width=0.45\textwidth]{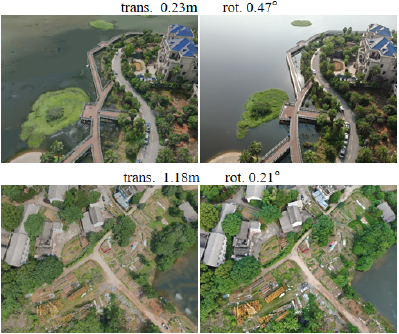}
    \caption{\textbf{Visual results of images rendered from the estimated poses.} The localization error for each pose is shown at the top of the image.}
    \label{eval_ori}
\end{figure} 

\subsection{Visual Localization}
\label{5.2}
\paragraph{Baselines.} We evaluate the performance of various keypoint descriptors, including SIFT~\cite{david2004distinctive} and Superpoint~\cite{detone2018superpoint}, as well as matchers such as Nearest Neighbor and the learnable SuperGlue~\cite{sarlin2020superglue}. We also compare the results with those obtained using the detector-free matcher LoFTR~\cite{sun2021loftr}. For pose estimation, we employ the PnP RANSAC solver, with and without gravity information.

\paragraph{Evaluation protocol.} Our evaluation follows the standard localization benchmark protocol~\cite{yan2022crossloc,sarlin2021back}, reporting results under the thresholds of $(1 m, 1^{\circ})$, $(3m, 3^{\circ})$ and $(5 m, 5^{\circ})$.

\begin{table*}[h!]
\centering
\small
\begin{tabular}{ccccccc}
\toprule
\multicolumn{1}{c||}{\multirow{2}{*}{Method}} & \multicolumn{3}{c|}{-}    & \multicolumn{3}{c}{Gravity-guided} \\ \cline{2-7} 
\multicolumn{1}{c||}{}                        & $(1m, 1^{\circ})$ & $(3m, 3^{\circ})$ & \multicolumn{1}{c|}{$(5m, 5^{\circ})$} & $(1m, 1^{\circ})$      & $(3m, 3^{\circ})$      & $(5m, 5^{\circ})$      \\ \hline
\multicolumn{1}{c||}{ LoFTR(Top @1) }    & 39.34  & 89.34  &           \multicolumn{1}{l|}{89.86}             &    43.18    &    91.78    &    93.01    \\ 
\multicolumn{1}{c||}{ LoFTR(Top @3) }    & 41.26  & 93.53  &           \multicolumn{1}{l|}{93.88}             &    43.18    &    93.88    &    94.58    \\ \hline
\multicolumn{1}{c||}{ SIFT + NN(Top @3) }    & 34.79  & 91.26 &         \multicolumn{1}{l|}{93.01}             &    34.09    &    90.73    &    91.78   \\ 
\multicolumn{1}{c||}{ SPP + NN(Top @3) }    & \textbf{41.96}  & 93.71  &           \multicolumn{1}{l|}{94.06}             &    43.71  &    94.93   &    95.45    \\ 
\multicolumn{1}{c||}{ SPP + SPG(Top @3) }    & 40.73  & \textbf{94.76}  &           \multicolumn{1}{l|}{\textbf{94.76}}             &    \textbf{44.58}    &    \textbf{96.15}    &    \textbf{96.68}    \\ 
\bottomrule     
\end{tabular}
\caption{\textbf{Visual localization results on UAVD4L.} We evaluate the localization performance of different feature detectors and descriptors using the two-stages UAV localization pipeline with \slash without sensor priors. Note that all methods use OpenIBL for image retrieval.}
\label{tab4.2}
\end{table*}

\paragraph{Results.} The quantitative results are presented in Table~\ref{tab4.2}. 
The Superpoint+Superglue outperforms other conterparts in all metric except $(1m,1^\circ)$. This may be attributed to the fact that query images of UAVD4L were taken under favorable lighting conditions and feature rich textures that facilitate keypoint detection by SuperPoint.
Additionally, we found that the gravity-guided PnP RANSAC method~\cite{yan2023long} enhances localization accuracy regardless of the matching algorithm used.
Qualitative results can be found in Figure~\ref{retrieval_label} and Figure~\ref{eval_ori}.
Figure~\ref{retrieval_label} provide the feature matching results, while Figure~\ref{eval_ori} provides a comparison between rendered estimated pose images and their corresponding original images, further demonstrating the precision of our results.

\subsection{Target Tracking}
\label{5.3}

To evaluate the target tracking performance of the proposed method, we capture people and car from two trajectories, and using the data recorded by RTK as their GT position. 

\paragraph{Evaluation protocol.} We adopt the translation difference between the ground truth 3D coordinates and the computed 3D coordinates to represent the target tracking estimation error.

\begin{figure*}[htbp]
\centering
    \includegraphics[width=0.90\textwidth]{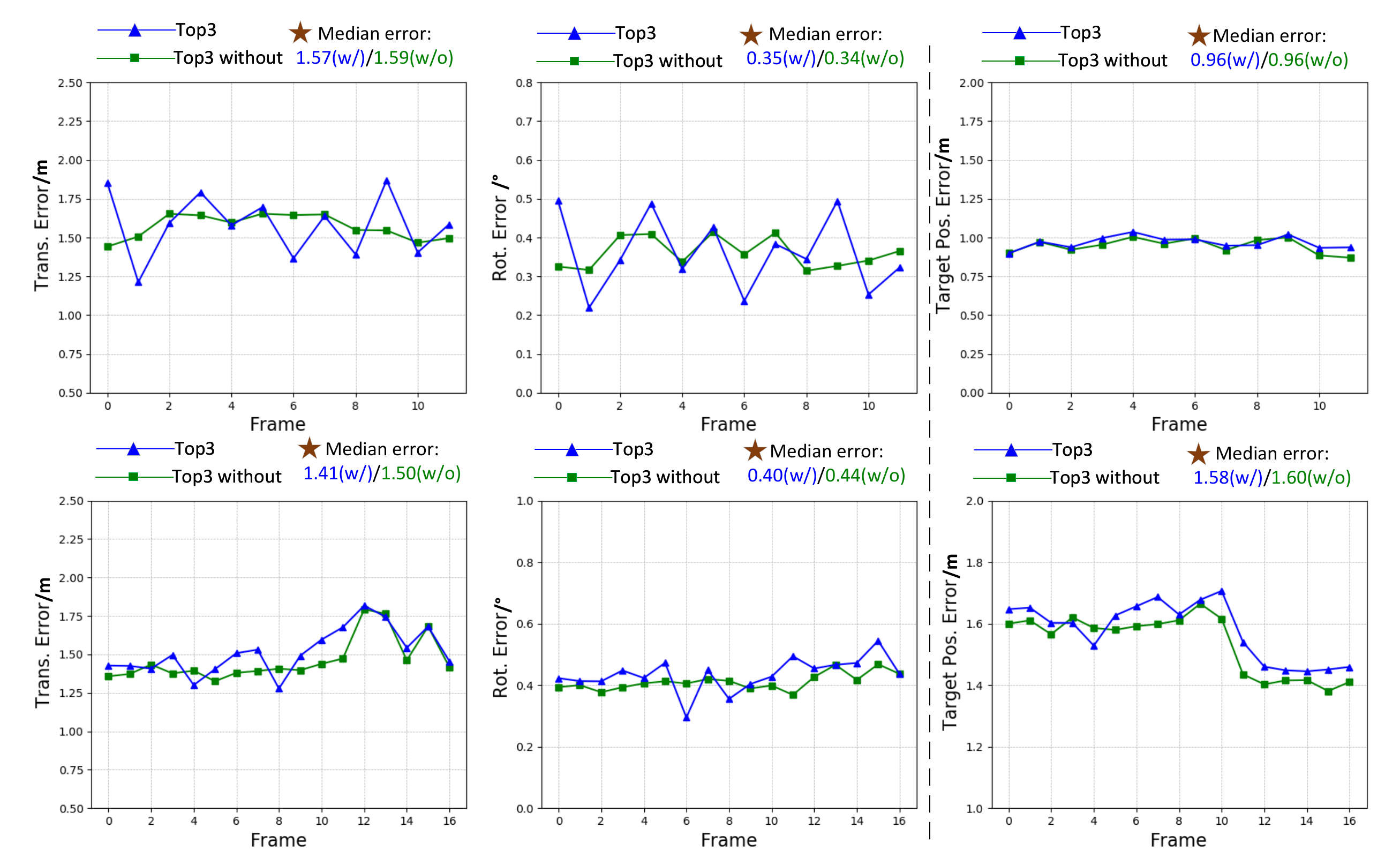}
    \caption{\textbf{Localization error results for different trajectory.} For two target tracking trajectories, we report the UAV localization and target tracking error with or without sensor prior. }
    \label{target_fig2}
\end{figure*}

\paragraph{Results.} Figure~\ref{target_fig2} shows two sequential UAV localization and target tracking results, with and without the use of sensor information for localization.

\begin{figure}[htbp]
    \centering
    \includegraphics[width=0.49\textwidth]{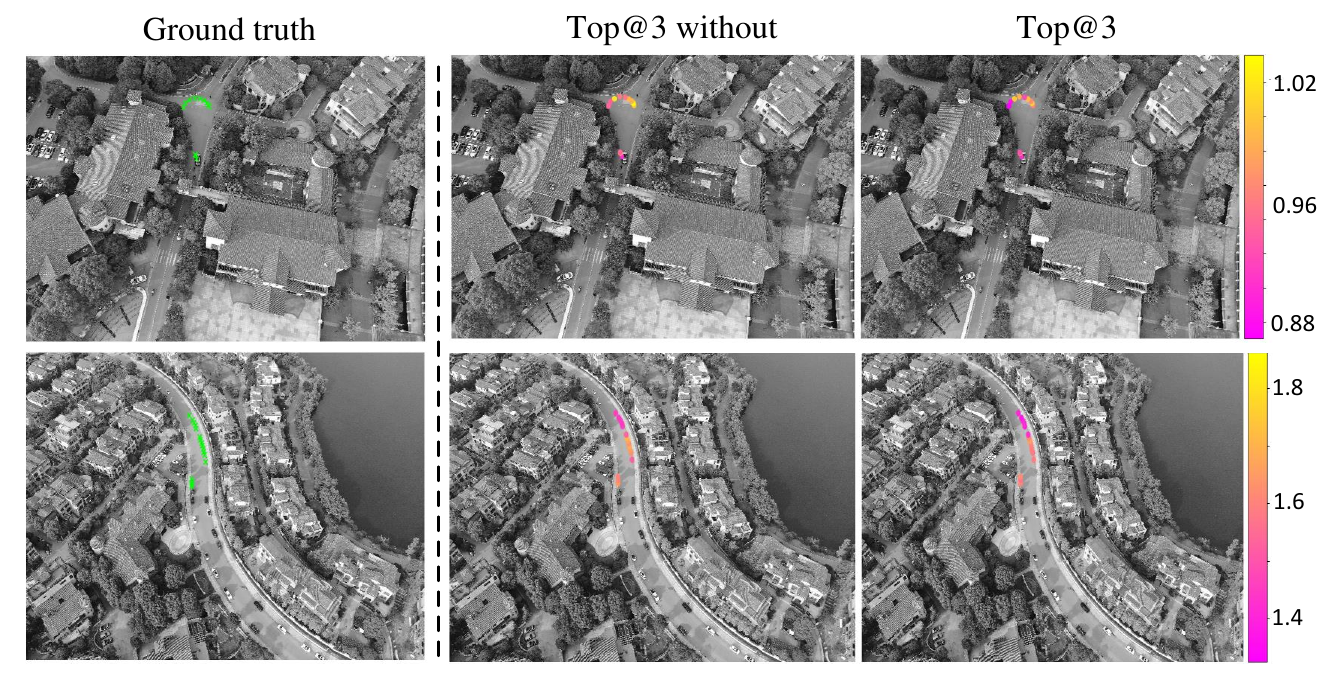}
    \caption{\textbf{Qualitative results of target tracking.} The left side of the figure displays the ground truth positions of the tracking target, represented by green points. The right side of the figure shows the estimated target position error, measured in meters as a translation error.}
    \label{target_fig}
\end{figure}

Due to the accurate UAV localization, the target position achieves meter-level accuracy. 
Figure~\ref{target_fig} visualizes the difference between the estimated target trajectory and the true trajectory under different UAV localization results. It can be seen that target estimation using sensor-guided localization results in a smaller position error.

\begin{table}[!ht]
\footnotesize 
\begin{tabular}{llll}
      \toprule
Hierarchical render ($m$,$^\circ$) & \multicolumn{3}{l}{$(1m, 1^{\circ})$ \slash $(3m, 3^{\circ})$ \slash $(5m, 5^{\circ})$} \\
\midrule
$H=150$, $\theta^{r}_{pitch}=45$ & \multicolumn{3}{l}{ 32.69 \slash 79.2 \slash 79.37} \\
$H=150$ \& $100$, $\theta^{r}_{pitch}=45$ & \multicolumn{3}{l}{ 42.63 \slash 89.86 \slash 90.03} \\
$H=150$ \& $100$, $\theta^{r}_{pitch}=45$ \& $0$ & \multicolumn{3}{l}{ \textbf{44.58} \slash \textbf{96.15} \slash \textbf{96.68}} \\
  \bottomrule
\end{tabular}
  \caption{\textbf{Ablation study.} The visual localization results of different hierarchical render setting are reported on the UAVD4L.} 
  \label{tab4.3}
\end{table}

\subsection{Ablation Studies}
\label{5.4}

To fully understand the impact of different hierarchical render settings on visual localization, we evaluate three render setting with results show in Table \ref{tab4.3}. Only render single level of altitude and angle obtains the lowest accuracy, while render multiple levels of attitude and angle attains the favorable accuracy. This indicates increasing render layer by altitude and angle improves the results. We conduct these experiment using the same image retrieval (OpenIBL), feature point detection and matching (SPP+SPG), and the gravity-guided PnP RANSAC on UAVD4L.

\section{Conclusion}
\label{conclusion}
We introduce a large-scale dataset, UAVD4L, for UAV 6-DoF localization in GPS-denied environments. The dataset contains a world-aligned textured 3D reference model and query images with accurate GT poses. However, generating an accurate GT poses of low illumination images is difficult, so in the future we will apply advanced techniques to expand our dataset. Additionally, we implement a novel two-stage 6-DoF localization pipeline that fully utilizes comprehensive data rendering and rotation sensor information. Furthermore, based on the 6-DoF pose estimator, we design a hierarchical system to track designated objects on the ground and output their 3D positions.  We anticipate that the UAVD4L dataset will serve as a valuable resource and benchmark for future research in this field.

\paragraph{Acknowledgements.} The authors would like to acknowledge the support from the Natural Science Foundation of Hunan Province of China (2020JJ5671). 
{
    \small
    \bibliographystyle{ieeenat_fullname}
    \bibliography{main}

\begin{thebibliography}{53}
\providecommand{\natexlab}[1]{#1}
\providecommand{\url}[1]{\texttt{#1}}
\expandafter\ifx\csname urlstyle\endcsname\relax
  \providecommand{\doi}[1]{doi: #1}\else
  \providecommand{\doi}{doi: \begingroup \urlstyle{rm}\Url}\fi

\bibitem[Al-Kaff et~al.(2018)Al-Kaff, Martin, Garcia, de~la Escalera, and
  Armingol]{al2018survey}
Abdulla Al-Kaff, David Martin, Fernando Garcia, Arturo de~la Escalera, and
  Jos{\'e}~Mar{\'\i}a Armingol.
\newblock Survey of computer vision algorithms and applications for unmanned
  aerial vehicles.
\newblock \emph{Expert Sys. Appl.}, 2018.

\bibitem[Arandjelovic et~al.(2016)Arandjelovic, Gronat, Torii, Pajdla, and
  Sivic]{arandjelovic2016netvlad}
Relja Arandjelovic, Petr Gronat, Akihiko Torii, Tomas Pajdla, and Josef Sivic.
\newblock Netvlad: Cnn architecture for weakly supervised place recognition.
\newblock In \emph{CVPR}, 2016.

\bibitem[Balamurugan et~al.(2016)Balamurugan, Valarmathi, and
  Naidu]{balamurugan2016survey}
Ganesan Balamurugan, J Valarmathi, and VPS Naidu.
\newblock Survey on uav navigation in gps denied environments.
\newblock In \emph{SCOPES}, 2016.

\bibitem[Barbarani et~al.(2023)Barbarani, Mostafa, Bayramov, Trivigno, Berton,
  Masone, and Caputo]{barbarani2023local}
Giovanni Barbarani, Mohamad Mostafa, Hajali Bayramov, Gabriele Trivigno,
  Gabriele Berton, Carlo Masone, and Barbara Caputo.
\newblock Are local features all you need for cross-domain visual place
  recognition?
\newblock In \emph{CVPR}, 2023.

\bibitem[Bejiga et~al.(2017)Bejiga, Zeggada, Nouffidj, and
  Melgani]{bejiga2017convolutional}
Mesay~Belete Bejiga, Abdallah Zeggada, Abdelhamid Nouffidj, and Farid Melgani.
\newblock A convolutional neural network approach for assisting avalanche
  search and rescue operations with uav imagery.
\newblock \emph{Remote Sens.}, 2017.

\bibitem[Bianchi and Barfoot(2021)]{bianchi2021uav}
Mollie Bianchi and Timothy~D Barfoot.
\newblock Uav localization using autoencoded satellite images.
\newblock \emph{IEEE Robot. Autom.}, 2021.

\bibitem[Cai et~al.(2022)Cai, Zhou, Zhang, Xia, Qiao, and Zhao]{cai2022review}
Yiming Cai, Yao Zhou, Hongwen Zhang, Yuli Xia, Peng Qiao, and Junsuo Zhao.
\newblock Review of target geo-location algorithms for aerial remote sensing
  cameras without control points.
\newblock \emph{Appl. Sci.}, 2022.

\bibitem[Chen et~al.(2021)Chen, Wu, Mueller, and Sreenath]{chen2021real}
Shuxiao Chen, Xiangyu Wu, Mark~W Mueller, and Koushil Sreenath.
\newblock Real-time geo-localization using satellite imagery and topography for
  unmanned aerial vehicles.
\newblock In \emph{IROS}, 2021.

\bibitem[Choi and Myung(2020)]{choi2020brm}
Junho Choi and Hyun Myung.
\newblock Brm localization: Uav localization in gnss-denied environments based
  on matching of numerical map and uav images.
\newblock In \emph{IROS}, 2020.

\bibitem[Cisneros et~al.(2022)Cisneros, Yin, Zhang, Choset, and
  Scherer]{cisneros2022alto}
Ivan Cisneros, Peng Yin, Ji Zhang, Howie Choset, and Sebastian Scherer.
\newblock Alto: A large-scale dataset for uav visual place recognition and
  localization.
\newblock \emph{arXiv preprint}, 2022.

\bibitem[Couturier and Akhloufi(2021)]{couturier2021review}
Andy Couturier and Moulay~A Akhloufi.
\newblock A review on absolute visual localization for uav.
\newblock \emph{Rob. Autom. Syst.}, 2021.

\bibitem[David(2004)]{david2004distinctive}
Lowe David.
\newblock Distinctive image features from scale-invariant keypoints.
\newblock \emph{IJCV}, 2004.

\bibitem[DeTone et~al.(2018)DeTone, Malisiewicz, and
  Rabinovich]{detone2018superpoint}
Daniel DeTone, Tomasz Malisiewicz, and Andrew Rabinovich.
\newblock Superpoint: Self-supervised interest point detection and description.
\newblock In \emph{CVPR}, 2018.

\bibitem[Ding et~al.(2023)Ding, Yang, Larsson, Olsson, and
  Åström]{haralick1994review}
Yaqing Ding, Jian Yang, Viktor Larsson, Carl Olsson, and Kalle Åström.
\newblock Revisiting the p3p problem.
\newblock 2023.

\bibitem[Fischler and Bolles(1981)]{fischler1981random}
Martin~A Fischler and Robert~C Bolles.
\newblock Random sample consensus: a paradigm for model fitting with
  applications to image analysis and automated cartography.
\newblock \emph{Commun. ACM}, 1981.

\bibitem[Ge et~al.(2020)Ge, Wang, Zhu, Zhao, and Li]{ge2020self}
Yixiao Ge, Haibo Wang, Feng Zhu, Rui Zhao, and Hongsheng Li.
\newblock Self-supervising fine-grained region similarities for large-scale
  image localization.
\newblock In \emph{ECCV}, 2020.

\bibitem[Goforth and Lucey(2019)]{goforth2019gps}
Hunter Goforth and Simon Lucey.
\newblock Gps-denied uav localization using pre-existing satellite imagery.
\newblock In \emph{ICRA}, 2019.

\bibitem[Gordo et~al.(2017)Gordo, Almazan, Revaud, and Larlus]{gordo2017end}
Albert Gordo, Jon Almazan, Jerome Revaud, and Diane Larlus.
\newblock End-to-end learning of deep visual representations for image
  retrieval.
\newblock \emph{IJCV}, 2017.

\bibitem[Guisado-Pintado et~al.(2019)Guisado-Pintado, Jackson, and
  Rogers]{guisado20193d}
Emilia Guisado-Pintado, Derek~WT Jackson, and David Rogers.
\newblock 3d mapping efficacy of a drone and terrestrial laser scanner over a
  temperate beach-dune zone.
\newblock \emph{Geomorphology}, 2019.

\bibitem[Gurgu et~al.(2022)Gurgu, Queralta, and Westerlund]{gurgu2022vision}
Marius-Mihail Gurgu, Jorge~Pe{\~n}a Queralta, and Tomi Westerlund.
\newblock Vision-based gnss-free localization for uavs in the wild.
\newblock In \emph{ICMERR}, 2022.

\bibitem[Huang et~al.(2020)Huang, Zhang, and Zhao]{huang2020high}
Chao Huang, Hongmei Zhang, and Jianhu Zhao.
\newblock High-efficiency determination of coastline by combination of tidal
  level and coastal zone dem from uav tilt photogrammetry.
\newblock \emph{Remote Sens.}, 2020.

\bibitem[Kanellakis and Nikolakopoulos(2017)]{kanellakis2017survey}
Christoforos Kanellakis and George Nikolakopoulos.
\newblock Survey on computer vision for uavs: Current developments and trends.
\newblock \emph{J. Intell. Rob. Syst. Theor. Appl.}, 2017.

\bibitem[Kinnari et~al.(2021)Kinnari, Verdoja, and Kyrki]{kinnari2021gnss}
Jouko Kinnari, Francesco Verdoja, and Ville Kyrki.
\newblock Gnss-denied geolocalization of uavs by visual matching of onboard
  camera images with orthophotos.
\newblock In \emph{ICRA}, 2021.

\bibitem[Kinnari et~al.(2022)Kinnari, Renzulli, Verdoja, and
  Kyrki]{kinnari2022lsvl}
Jouko Kinnari, Riccardo Renzulli, Francesco Verdoja, and Ville Kyrki.
\newblock Lsvl: Large-scale season-invariant visual localization for uavs.
\newblock \emph{arXiv preprint}, 2022.

\bibitem[Larson et~al.(2021)Larson, Ianuzzi, Oliva, Lorence, Sheffer,
  et~al.]{larson2021autonomous}
Edward Larson, Arthur Ianuzzi, Renier Oliva, Dennis Lorence, Graham Sheffer,
  et~al.
\newblock Autonomous underwater survey apparatus and system, 2021.
\newblock US Patent 11,072,405.

\bibitem[Lepetit et~al.(2009)Lepetit, Moreno-Noguer, and Fua]{lepetit2009ep}
Vincent Lepetit, Francesc Moreno-Noguer, and Pascal Fua.
\newblock Epnp: An accurate o(n) solution to the pnp problem.
\newblock \emph{IJCV}, 2009.

\bibitem[Liu et~al.(2023)Liu, Bai, Wang, Wu, Sun, Guo, and Geng]{liu2023uav}
Yu Liu, Jing Bai, Gang Wang, Xiaobo Wu, Fangde Sun, Zhengqiang Guo, and Hujun
  Geng.
\newblock Uav localization in low-altitude gnss-denied environments based on
  poi and store signage text matching in uav images.
\newblock \emph{Drones}, 2023.

\bibitem[Lu et~al.(2018)Lu, Xue, Xia, and Zhang]{lu2018survey}
Yuncheng Lu, Zhucun Xue, Gui-Song Xia, and Liangpei Zhang.
\newblock A survey on vision-based uav navigation.
\newblock \emph{Geo-Spatial Inf. Sci.}, 2018.

\bibitem[Marcu et~al.(2018)Marcu, Costea, Slusanschi, and
  Leordeanu]{marcu2018multi}
Alina Marcu, Dragos Costea, Emil Slusanschi, and Marius Leordeanu.
\newblock A multi-stage multi-task neural network for aerial scene
  interpretation and geolocalization.
\newblock \emph{arXiv preprint}, 2018.

\bibitem[Moreau et~al.(2022)Moreau, Piasco, Tsishkou, Stanciulescu, and
  de~La~Fortelle]{moreau2022lens}
Arthur Moreau, Nathan Piasco, Dzmitry Tsishkou, Bogdan Stanciulescu, and Arnaud
  de La~Fortelle.
\newblock Lens: Localization enhanced by nerf synthesis.
\newblock In \emph{Proc. Mach. Learn. Res.}, 2022.

\bibitem[Mughal et~al.(2021)Mughal, Khokhar, and Shahzad]{mughal2021assisting}
Muhammad~Hamza Mughal, Muhammad~Jawad Khokhar, and Muhammad Shahzad.
\newblock Assisting uav localization via deep contextual image matching.
\newblock \emph{IEEE J. Sel. Top. Appl. Earth Obs. Remote Sens.}, 2021.

\bibitem[Nassar and ElHelw(2018)]{nassar2020aerial}
Ahmed Nassar and Mohamed ElHelw.
\newblock Aerial imagery registration using deep learning for uav
  geolocalization.
\newblock In \emph{CVPRW}, 2018.

\bibitem[Nath et~al.(2022)Nath, Cheng, and Behzadan]{nath2022drone}
Nipun~D Nath, Chih-Shen Cheng, and Amir~H Behzadan.
\newblock Drone mapping of damage information in gps-denied disaster sites.
\newblock \emph{Adv. Eng. Inf.}, 2022.

\bibitem[Panek et~al.(2022)Panek, Kukelova, and Sattler]{panek2022meshloc}
Vojtech Panek, Zuzana Kukelova, and Torsten Sattler.
\newblock Meshloc: Mesh-based visual localization.
\newblock In \emph{ECCV}, 2022.

\bibitem[Patel et~al.(2020)Patel, Barfoot, and Schoellig]{patel2020visual}
Bhavit Patel, Timothy~D Barfoot, and Angela~P Schoellig.
\newblock Visual localization with google earth images for robust global pose
  estimation of uavs.
\newblock In \emph{ICRA}, 2020.

\bibitem[Sarlin et~al.(2019)Sarlin, Cadena, Siegwart, and
  Dymczyk]{sarlin2019coarse}
Paul-Edouard Sarlin, Cesar Cadena, Roland Siegwart, and Marcin Dymczyk.
\newblock From coarse to fine: Robust hierarchical localization at large scale.
\newblock In \emph{CVPR}, 2019.

\bibitem[Sarlin et~al.(2020)Sarlin, DeTone, Malisiewicz, and
  Rabinovich]{sarlin2020superglue}
Paul-Edouard Sarlin, Daniel DeTone, Tomasz Malisiewicz, and Andrew Rabinovich.
\newblock Superglue: Learning feature matching with graph neural networks.
\newblock In \emph{CVPR}, 2020.

\bibitem[Sarlin et~al.(2021)Sarlin, Unagar, Larsson, Germain, Toft, Larsson,
  Pollefeys, Lepetit, Hammarstrand, Kahl, et~al.]{sarlin2021back}
Paul-Edouard Sarlin, Ajaykumar Unagar, Mans Larsson, Hugo Germain, Carl Toft,
  Viktor Larsson, Marc Pollefeys, Vincent Lepetit, Lars Hammarstrand, Fredrik
  Kahl, et~al.
\newblock Back to the feature: Learning robust camera localization from pixels
  to pose.
\newblock In \emph{CVPR}, 2021.

\bibitem[Sarlin et~al.(2022)Sarlin, Dusmanu, Sch{\"o}nberger, Speciale, Gruber,
  Larsson, Miksik, and Pollefeys]{sarlin2022lamar}
Paul-Edouard Sarlin, Mihai Dusmanu, Johannes~L Sch{\"o}nberger, Pablo Speciale,
  Lukas Gruber, Viktor Larsson, Ondrej Miksik, and Marc Pollefeys.
\newblock Lamar: Benchmarking localization and mapping for augmented reality.
\newblock In \emph{ECCV}, 2022.

\bibitem[Shetty and Gao(2019)]{shetty2019uav}
Akshay Shetty and Grace~Xingxin Gao.
\newblock Uav pose estimation using cross-view geolocalization with satellite
  imagery.
\newblock In \emph{ICRA}, 2019.

\bibitem[Silvagni et~al.(2017)Silvagni, Tonoli, Zenerino, and
  Chiaberge]{silvagni2017multipurpose}
Mario Silvagni, Andrea Tonoli, Enrico Zenerino, and Marcello Chiaberge.
\newblock Multipurpose uav for search and rescue operations in mountain
  avalanche events.
\newblock \emph{Geomatics Nat. Hazards Risk}, 2017.

\bibitem[Sun et~al.(2021)Sun, Shen, Wang, Bao, and Zhou]{sun2021loftr}
Jiaming Sun, Zehong Shen, Yuang Wang, Hujun Bao, and Xiaowei Zhou.
\newblock Loftr: Detector-free local feature matching with transformers.
\newblock In \emph{CVPR}, 2021.

\bibitem[Taira et~al.(2018)Taira, Okutomi, Sattler, Cimpoi, Pollefeys, Sivic,
  Pajdla, and Torii]{taira2018inloc}
Hajime Taira, Masatoshi Okutomi, Torsten Sattler, Mircea Cimpoi, Marc
  Pollefeys, Josef Sivic, Tomas Pajdla, and Akihiko Torii.
\newblock Inloc: Indoor visual localization with dense matching and view
  synthesis.
\newblock In \emph{CVPR}, 2018.

\bibitem[Wu et~al.(2021)Wu, Du, Chen, and Jing]{wu2021coarse}
Songbing Wu, Chun Du, Hao Chen, and Ning Jing.
\newblock Coarse-to-fine uav image geo-localization using multi-stage
  lucas-kanade networks.
\newblock In \emph{ICTC}, 2021.

\bibitem[Xu et~al.(2018)Xu, Pan, Du, Li, Jing, and Wu]{xu2018vision}
Yingxiao Xu, Long Pan, Chun Du, Jun Li, Ning Jing, and Jiangjiang Wu.
\newblock Vision-based uavs aerial image localization: A survey.
\newblock In \emph{SIGSPATIAL}, 2018.

\bibitem[Xu et~al.(2022)Xu, Wu, Du, Li, and Jing]{xu2022uav}
Yingxiao Xu, Songbing Wu, Chun Du, Jun Li, and Ning Jing.
\newblock Uav image geo-localization by point-line-patch feature matching and
  iclk optimization.
\newblock In \emph{Int. Conf. Geoinformatics}, 2022.

\bibitem[Yan et~al.(2022)Yan, Zheng, Reding, Li, and
  Doytchinov]{yan2022crossloc}
Qi Yan, Jianhao Zheng, Simon Reding, Shanci Li, and Iordan Doytchinov.
\newblock Crossloc: Scalable aerial localization assisted by multimodal
  synthetic data.
\newblock In \emph{CVPR}, 2022.

\bibitem[Yan et~al.(2023)Yan, Liu, Wang, Shen, Peng, Liu, Zhang, Zhang, and
  Zhou]{yan2023long}
Shen Yan, Yu Liu, Long Wang, Zehong Shen, Zhen Peng, Haomin Liu, Maojun Zhang,
  Guofeng Zhang, and Xiaowei Zhou.
\newblock Long-term visual localization with mobile sensors.
\newblock In \emph{CVPR}, 2023.

\bibitem[Yin et~al.(2021)Yin, Xu, Zhang, Choset, and Scherer]{yin2021i3dloc}
Peng Yin, Lingyun Xu, Ji Zhang, Howie Choset, and Sebastian Scherer.
\newblock i3dloc: Image-to-range cross-domain localization robust to
  inconsistent environmental conditions.
\newblock \emph{arXiv preprint}, 2021.

\bibitem[Yin et~al.(2023)Yin, Cisneros, Zhao, Zhang, Choset, and
  Scherer]{yin2023isimloc}
Peng Yin, Ivan Cisneros, Shiqi Zhao, Ji Zhang, Howie Choset, and Sebastian
  Scherer.
\newblock isimloc: Visual global localization for previously unseen
  environments with simulated images.
\newblock \emph{IEEE Trans. Rob.}, 2023.

\bibitem[Zhang et~al.(2022)Zhang, Tang, Qiu, Huang, Fang, Cui, Dong, Zhu, and
  Tan]{zhang2022rendernet}
Jiahui Zhang, Shitao Tang, Kejie Qiu, Rui Huang, Chuan Fang, Le Cui, Zilong
  Dong, Siyu Zhu, and Ping Tan.
\newblock Rendernet: Visual relocalization using virtual viewpoints in
  large-scale indoor environments.
\newblock \emph{arXiv preprint}, 2022.

\bibitem[Zhang et~al.(2021)Zhang, Sattler, and Scaramuzza]{zhang2021reference}
Zichao Zhang, Torsten Sattler, and Davide Scaramuzza.
\newblock Reference pose generation for long-term visual localization via
  learned features and view synthesis.
\newblock \emph{IJCV}, 2021.

\bibitem[Zheng et~al.(2020)Zheng, Wei, and Yang]{zheng2020university}
Zhedong Zheng, Yunchao Wei, and Yi Yang.
\newblock University-1652: A multi-view multi-source benchmark for drone-based
  geo-localization.
\newblock In \emph{MM}, 2020.

\end{thebibliography}
}
\end{document}